\documentclass[conference]{IEEEtran}

\usepackage{cite}
\usepackage{amsmath,amssymb,amsfonts}
\usepackage{graphicx}
\usepackage{textcomp}
\usepackage{cleveref}
\usepackage{float}
\usepackage{soul}
\usepackage{color,subcaption}
\emergencystretch=\maxdimen
\usepackage{url}
\usepackage{booktabs}

\def\BibTeX{{\rm B\kern-.05em{\sc i\kern-.025em b}\kern-.08em
    T\kern-.1667em\lower.7ex\hbox{E}\kern-.125emX}}

\usepackage{subcaption}
\DeclareCaptionLabelSeparator{emdash}{\textemdash}

\begin{document}

\title{CBR - Boosting Adaptive Classification By Retrieval of Encrypted Network Traffic with Out-of-distribution }

\author{\IEEEauthorblockN{Amir Lukach,  Ran Dubin, Amit Dvir}
\IEEEauthorblockA{\textit{Department of Computer Science} \\ \textit{Ariel Cyber Innovation Center}\\ Ariel University, Israel\\
amirluckach@gmail.com, rand, amitdv@ariel.ac.il}
\and
\IEEEauthorblockN{Chen Hajaj}
\IEEEauthorblockA{\textit{Department of Industrial Engineering and Management} \\\textit{Data Science and Artificial Intelligence Research Center}\\ Ariel University, Israel\\
chenha@ariel.ac.il}
}

\maketitle

\begin{abstract}
Encrypted network traffic Classification tackles the problem from different approaches and with different goals. One of the common approaches is using Machine learning or Deep Learning-based solutions on a fixed number of classes, leading to misclassification when an unknown class is given as input. One of the solutions for handling unknown classes is to retrain the model, however, retraining models every time they become obsolete is both resource and time-consuming. Therefore, there is a growing need to allow classification models to detect and adapt to new classes dynamically, without retraining, but instead able to detect new classes using few shots learning\cite{wang2020generalizing}. In this paper, we introduce Adaptive Classification By Retrieval \textit{CBR}, a novel approach for encrypted network traffic classification. Our new approach is based on an ANN-based method, which allows us to effectively identify new and existing classes without retraining the model. The novel approach is simple, yet effective and achieved similar results to RF with up to 5\% difference (usually less than that) in the classification tasks while having a slight decrease in the case of new samples (from new classes) without retraining. To summarize, the new method is a real-time classification, which can classify new classes without retraining. Furthermore, our solution can be used as a complementary solution alongside RF or any other machine/deep learning classification method, as an aggregated solution.
\end{abstract}


\maketitle

\section{Introduction}
\label{sec:introduction}
Classical Machine Learning (ML) and Deep Learning (DL) models are applicable in the scope of encrypted network traffic classification\cite{nlp,goodman2020packet2vec,li2020weighted,corizzo2020feature,bar2022simcse, flowpic2021, deepmal,horowicz2022few}, and thus served as the baseline for classifying encrypted traffic. Due to recent massive changes in the internet protocols, e.g., QUIC \cite{quic}, HTTP/3, TLS 1.3 and DoH \cite{bottger2019empirical}, Deep Packet Inspection (DPI) traditional classification methods, which leverage DNS and Service Name Indicator (SNI), to identify encrypted network traffic, will soon no longer be usable. Therefore, advanced encrypted traffic flow classification algorithms are needed  \cite{flowclassification}. Recently, the classification models adopted various data representations, including a growing number of works that have used Natural Language Processing (NLP) \cite{nlp} techniques, such as transforming the flow into a language to use word embedding\cite{goodman2020packet2vec,li2020weighted,corizzo2020feature,bar2022simcse}, while others have converted the network flow into an image to harness image processing techniques and equivalent DL architectures \cite{ flowpic2021, WeiWangMalwareTrafficClassification, deepmal,horowicz2022few}. 

In order to validate that an encrypted network traffic classification model will efficiently classify unknown samples correctly, there are a few requirements \cite{aws_update_neural_network}: \textbf{I.} Acquire a large amount of training data for the new class, \textbf{II.} Add it to the dataset, which was initially used to train the classifier, and \textbf{III.} Retrain the classifier on the combined data set. Regarding ML (and especially DL) models, every time they become obsolete is both resource and time-consuming, especially when the application is complex. Therefore, there is a growing need to allow classification models to detect and adapt to new classes dynamically, without retraining\cite{schulz2020extending}. One way of allowing the above is using Nearest neighbor search, one of the most well-known tools in many research areas \cite{korn1998fast,bustos2006graphics}. In some cases, a generic nearest neighbor search under a suitable distance or measure of similarity offers dramatic quality improvements \cite{beatenpath}. K-Nearest Neighbors algorithm (KNN), by a plurality of its neighbors, is where the output is the class's label, and $k$ is a positive (usually small) integer. To obtain efficient algorithms, one may use ANN \cite{aumuller2020ann} in which the returned neighbors may be an approximation of the true nearest neighbors. Usually, this means that the answer to finding the nearest neighbors to a query point is judged by the distance of the query point to the set of its true nearest neighbors. ANN \cite{ANN_ALGORITHM} is a variation of KNN that aims to limit the number of training samples that each new test point is compared with before returning a result. 

In this paper, we present a vector-based data representation that classifies encrypted network traffic and can identify and add new classes without the need for retraining. Our main motivtion was to find a balanced solution, the ability to easily learn new classes, that provides feasible and accurate results for different types of encrypted network traffic classification challenges.
To achieve this goal, our approach relies on the ANN method, which excels in quickly matching a new vector against a collection of already labeled and indexed vectors. This matching process is based on the principle of Euclidean similarity distance (there are other distance metrics, such as Cosine distance)\cite{distmetric}. First, it performs a fast real-time classification; second, it allows users and security software to accurately identify new classes without the need for retraining the mode.
\iffalse
\hl{A: is it not the same as the last paragraph?} Our encrypted network traffic classification system (named \textit{CBR}) is based on the ANN model, which enables us to deal with new classes. CBR is a novel approach for encrypted network classification that tackles the disadvantage of retraining by using a vector search instead, which allows us to effectively identify new and existing classes by a simple yet effective method. 
\fi
Our novel approach is based on the idea that if a sample distance to other samples is higher than a defined threshold, we classify it as a new class. By that, we omit the necessity to retrain the model if new classes or instances are introduced, as the vector search process is dynamic. This way, our system can extend itself using only a few samples from a new class. Note that our proposed solution uses only statistical features, so it should be robust to future planned protocol changes. 
To summarize, the contributions of this 
paper are as follows:
\begin{enumerate}
\item To the best of our knowledge, this is the first paper that suggests encrypted network classification based on ANN that handles few shots learning and OOD.
\item Identification of network anomalies for information security purposes, by classifying them as new classes.
\item Fast detection and addition of new classes without retraining using a real-time classification model that needs only a few shots from the new classes, while still providing accuracy that is comparable to standard ML models (e.g., RF). Note, that when combining OOD to our solution, in some cases it even outperformed RF. In addition, it can also be used with RF as an aggregated solution.
\item Preparing a classification solution that should comply with fully encrypted networks.
\end{enumerate}

\iffalse
\begin{figure}[ht]
    \subfloat[ANN with OOD]{\label{fig_ood_with_ann}\includegraphics[width=.8\linewidth]{images/annwithood.jpg}} \\
    \subfloat[ANN without OOD]{\label{fig_ann_without_ood}\includegraphics[width=.8\linewidth]{images/annwithoutood.jpg}}
    \caption{ANN with and without OOD.}
    \label{fig_ood_with_and_without_ann}
\end{figure}
\begin{figure}[ht]
\centering
\includegraphics[width=.5\linewidth]{images/annwithood.jpg}
\caption{Combining Out-of-distribution detection with ANN}{}
\label{fig_ood_with_ann}
\end{figure}
\begin{figure}[ht]
\centering
\includegraphics[width=.8\linewidth]{images/annwithoutood.jpg}
\caption{ANN without OOD}{}
\label{fig_ann_without_ood}
\end{figure}
\fi
\section{Related work}
In recent years, DL models have become the prominent method for network traffic classification~\cite{weiwangmalwaretrafficclassificationm2cnn2017, wang1dcnn, citemtatls, cite5}. 
The DL's work has multiple scopes and domains such as classification \cite{boa_conf, cite6, conti_new_2017, PINHEIRO20198}. Some works have converted the network flow into an image to harness image processing techniques and equivalent DL architectures \cite{ cite3, cite6, cite10, flowpic2021, WeiWangMalwareTrafficClassification, deepmal}, while others used NLP \cite{bar2022simcse} or graph neural networks \cite{CGNN}. In the cyber domain, works tackle the task of malware network traffic detection (benign, malicious) and classification \cite{MTAKDD19, deepmal, yesml2017, cite30, citemtatls, D201950}. In addition, some previous works suggest using classification by retrieval to solve specific video traffic analysis challenges, like motion detection, Chan et al.\cite{1505198}, and audio classification and segmentation, which allows real-time audio classification into basic types, Zhang et al.\cite{10.1117/12.325703}, however, these are very domain-specific solutions.
Although DL's architectures started to replace ML for traffic classification, in \cite{LICHY2023103000} the authors presented a comparative experiment between ML/DL algorithms that shows, in some cases, that ML algorithms such as RF \cite{RF} are more than enough. All the above works tackle the classification problem without discussing the challenge of retraining the model, which is needed due to a changing network traffic landscape.
Retraining a machine learning model involves updating the model to accommodate the new knowledge, which is necessary to perform well, most of the works use datasets with only a few classes, e.g. \cite{flowpic2021,chen2019all}, they use per-flow features and do not consider scenarios where new applications are progressively added to models. Therefore, the focus of these works is only on the problem of creating the most accurate classifier given immutable data for both the number of classes and the data for each class. These systems are based on creating a new training set and training a new model from scratch. It may, however, be inefficient and require high computation performance, to update them with new classes' classification.
The K-Nearest Neighbors (KNN) is one kind of lazy classification algorithm without the process of classifier training. By learning-to-hash algorithms, the KNN-based classification can be mapped to the hash table searching whose execution time and memory cost are both acceptable. Qi et al.\cite{QIBLOCKCHAIN} presented lightweight IoT traffic classification based on KNN \cite{KNN_ALGORITHM}.  Ma et al. \cite{improved_knn2020} proposed a method, based on the K-nearest neighbor (KNN) algorithm, which only needs a small amount of data to train a model. Moreover, the authors also presented a three-layer classification framework for encrypted network flows. ANN \cite{ANN_ALGORITHM} is a variation of KNN that aims to limit the training sample number that each new test point is compared with, before returning a result. Many efficient ANN implementations have been developed with diverse approaches, such as dimension reduction \cite{kramer2013dimensionality}, locality-sensitive hashing \cite{huang2015query}, and compressed sensing \cite{aumuller2020ann}, but none of them use the advantages of the ANN in the field of encrypted network classification.

\begin{table*}[t]
\begin{tabular}{|l|l|l|l|}\hline
\textbf{Feature} & \textbf{Slots in vector} & \textbf{Description} & \textbf{Used In} \\\hline
Bits per peak                & 3               & Summary of bits of every "peak" in data coming from destination to source &   \\\hline
First packets sizes          & 30              & First 30 packet sizes with sign (+ or -) by direction of communication    &   \\\hline
Beaconing    & 20 & Sum of packet size where the source is more active than the destination in 5 seconds windows & BOA \& MTA\\\hline
Bandwidth                     & 20              & Minimum and maximum delta of TCP window size in 5 seconds windows &            \\\hline
Statistics of packet sizes  & 4               & Min, max, mean, STD of packet sizesa&     BOA \& MTA   \\\hline
Size's delta between packets & 2               & Mean and STD deltas                                &      BOA \& MTA   \\\hline
Packets per second           & 2               & Packets per second forward and backward                               &       \\\hline
Inter-arrival time           & 9               & Min, max, and mean of bidirectional, forward, and backward              &       \\\hline
Silence windows              & 1               & Amount of silence windows longer than 1 second or 10 ms every 1 or 0.5 seconds & \\\hline
Amount of ACK packets        & 1               & Bidirectional count of TCP packets which contain ACK flag                                                               &   \\\hline
Big requests & 1  & Amount of client-to-server messages bigger than 200-byte ending session with a new request &  \\\hline 
Wavelet & 90  & FFT coefficient on first 90 packet sizes received &  BOA \& MTA\\\hline 
\end{tabular}
\caption{Feature sets.This table describes the features vector we used, the size in the vector of each of the features, and for which datasets we used it.}
\label{tab_features}
\end{table*}
\iffalse
\begin{figure*}[!]
    \centering
    \begin{subfigure}[t]{0.29\textwidth}
        \centering
        \includegraphics[height=1.8in]{images/flow.jpg}
        \caption{CBR Work Flow. Our solution starts with extracting features to CSV files, then splitting data to test and training sets, adding the training set to the elastic database, and then testing each test's PCAP sample by querying the elastic database.}
        \label{fig_acbr_methodology}
    \end{subfigure}%
    ~ 
    \begin{subfigure}[t]{0.29\textwidth}
        \centering
        \includegraphics[height=1.8in]{images/Flowtrain.jpg}
        \caption{CBR Training Flow.  The training flow is as follows: samples are split to test and train, features are extracted from each sample, where training data is added to elastic, and test data is saved separately.}
        \label{fig:trainingflow}
    \end{subfigure}
    ~
    \begin{subfigure}[t]{0.29\textwidth}
        \centering
        \includegraphics[height=1.8in]{images/Flownewclass.jpg}
    \caption{CBR Prediction/Inference Flow. The prediction/inference test is as follows: each sample is parsed, and compared to Elastic's query, and if the distance is equal or greater to a threshold then they are classified as a new class, else they are classified as one of the existing classes.}
    \label{fig:precitionflow}
    \end{subfigure}
    \caption{CBR Flows}
\end{figure*}
\fi
\section{\label{sec:Methodology}Methodology}
In this section, we will explain how our solution (CBR) tackles the problem of retraining, while still providing a fast and accurate classification.
Our goal is to create an adaptive, encrypted traffic classifier, which will be able to accurately identify existing, and new classes in a fast, real-time, and dynamic method. To achieve fast classification, we have chosen ANN, and have used a distributed search engine for storing and retrieving our samples, since it supports real-time search performance \cite{elastic_search} (note that our solution also supports using any other ANN search algorithm).
Moreover, our solution has the following benefits: near real-time operations, such as reading or writing data usually take less than a second to complete, and high performance. Using the distributed nature of the search engine enables it to process large volumes of data in parallel, and find the best matches for our queries \cite{elastic_search}.

The CBR's architecture steps are as follows (as can be seen in Figure \ref{fig_acbr_methodology}): The first step is to read the network data, afterwards we organize the data by connections 5-tuple (source IP, destination IP, protocol, source port, and destination port), from each sample, we extract the required features (see Table \ref{tab_features}) and normalize them, perform feature selection to use the minimal amount of features, to provide the fastest solution, use Random Forest as one of the inputs for classifying the classes, perform semi-supervised learning to find OOD and use it as input for CBR, use CBR to classify new, and existing classes and OOD. Then, inputs from RF and CBR are combined in order to classify current classes, new classes, and OOD (for new classes and OOD we use input from CBR, and use RF for the rest of the classification). 
In the training phase, to build our vector search database, we split the data into a training set (70\%) and a test set (30\%), and add the training set features' to the Elastic search database \cite{elastic_search}. For each sample from the test set, the model predicts the label. The inference phase of CBR is done by ANN queries from the database. Each ANN query selects K's closest samples, and we choose the class with the most hits.
\iffalse In the training part, each dataset comprises traffic samples (e.g., PCAP files), where each sample represents a traffic flow. From each sample, we extract the required features (see Table \ref{tab_features}) and normalize them. 
\fi 

\begin{figure}[H]
\centering
\includegraphics[height=4cm,width=8cm]{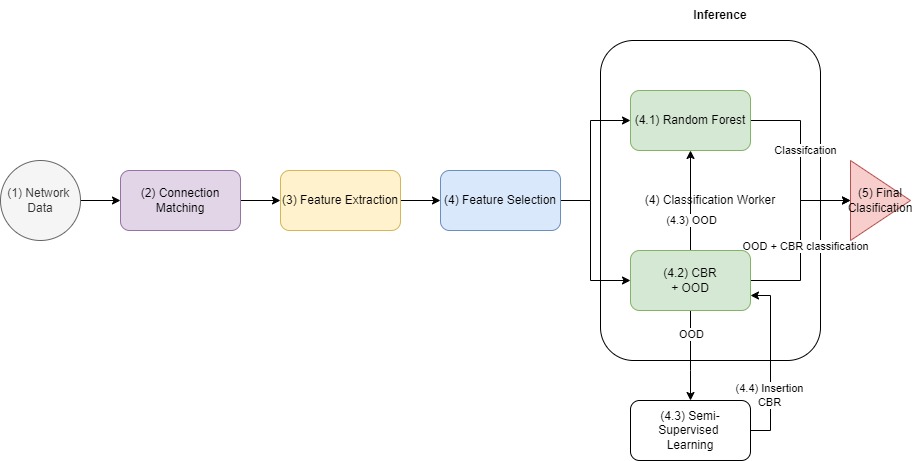}
\caption{CBR Architecture}{}
\label{fig_acbr_methodology}
\end{figure}
To summarize, in the training phase, each sample is transformed into a vector with a label. The model saves each training sample vector index and uses the label to map it to a class. The inference phase of CBR is responsible for labeling each test sample with the matching class by using the distance from the test sample to get $k$ samples from the database and then using the label of the sample with the max hits to decide on the label. Furthermore, if the distance between the current tested vector features to the closest class is larger than a defined threshold, we add a new class. In our case, we use the Euclidean distance metric to query the elastic search for K closest samples and select the class with the maximum hits' score. 
Furthermore, to prevent adding new classes from Out Of Distribution samples, we have used semi-supervised learning (we manually labeled some of the OOD) for Out-Of-Distribution detection, for samples, which are very distant from their closest class and have used existing dataset's features, without calculating features from related packet capture files. ANN methods have been used in the past for OOD detection, e.g. Sun et al.\cite{pmlr-v162-sun22d}, but in our case, we compared RF, VS ANN with OOD which was used to improve the classification of known classes and not just to detect the OOD. If we look at ANN without OOD, then, in this case, every test sample is labeled with the closest class, no matter how distant they are from the closest class.

The latter is displayed in Figure \ref{fig_ood_with_and_without_ann}. Moreover, there is also an option to use our solution, alongside RF as part of the same prediction flow, in which our solution can be used to remove OOD and combine with the results we receive from RF, as depicted in Figure \ref{fig_acbr_methodology}.
\iffalse
\begin{figure}[ht]
\centering
\includegraphics[height=8cm,width=8cm]{images/RF.jpg}
\caption{RF Architecture. Notice, that RF does not know how to distinguish OOD from known classes, so each OOD will be classified as one of the existing classes.}{}
\label{fig_rf}
\end{figure}
\fi
\begin{figure}
    \centering
    \includegraphics[width=.8\linewidth]{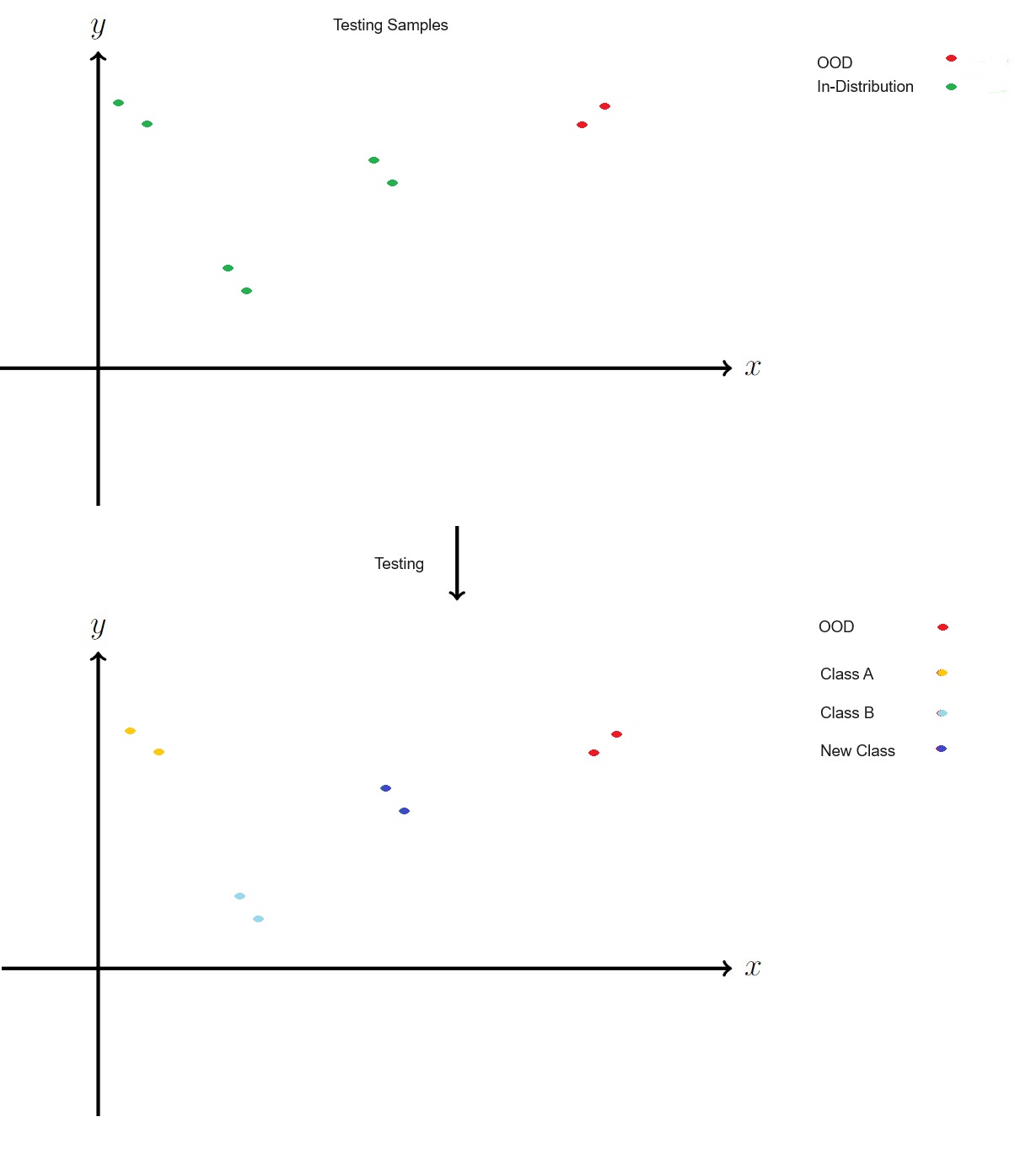}
    \caption{ANN with OOD}
    \label{fig_ood_with_and_without_ann}
\end{figure}
\iffalse
\begin{figure}[ht]
\centering
\includegraphics[height=16cm,width=8cm]{images/RF_Extended.jpg}
\caption{RF with CBR. Here, our vision is to integrate CBR and RF solutions, as an aggregated solution, where we start with CBR to remove OOD samples, and the output without OOD is used as input for the RF classifier. This way, we remove the false positive classification of OOD samples.}{}
\label{fig_rf_with_ood}
\end{figure}
\fi
%\begin{figure}[H]
%    \centering
%    \includegraphics[width=.8\linewidth]{images/Flowtrain.jpg}
%    \caption{One Shot Training Flow}
%    \label{fig:trainingflow}
%\end{figure}
%\begin{figure}[H]
%    \centering
%    \includegraphics[width=.8\linewidth]{images/Flownewclass.jpg}
%    \caption{One Shot Prediction Flow}
%    \label{fig:precitionflow}
%\end{figure}

\section{Experimental Design}
\label{sec:ExperimentalDesign}
We conducted a set of experiments to evaluate the effectiveness of the ANN-based approach for a set of classification tasks on encrypted network traffic from multiple known public datasets. Specifically, we looked at both malicious and benign datasets. Using the malicious datasets, we classified the malware family. For the benign datasets, we evaluated the Operation System (OS), browser classification, and the application's classification. The goal of our evaluation was, first to assess the slight decrease in our classifier's performance due to learning new classes, while classifying existing classes when comparing them to a classical machine learning classifier (e.g., RF). Second, to show the robustness of our ANN-based classifier in the classification of new classes.

\subsection{Datasets}
We used two common datasets for our evaluations: BOA\cite{boa_conf}, and MTA\cite{MTA}. 
The \textbf{BOA} dataset was presented in \cite{boa_conf} where the authors collected the data for more than two months in their lab, using a selenium web crawler for browser traffic. The dataset contains applications, such as YouTube and Facebook, labeled as browser traffic, and Dropbox and TeamViewer, labeled as non-browser traffic. The dataset contains more than 20,000 sessions. This dataset contains information on browsers, OSs, and applications with and without their correlated browser. The average duration of a session was 518 seconds, and on average, each session had 520 forward packets (the average forward traffic size was 261K bytes) and 637 backward packets (the average backward traffic size was 615K bytes). Almost all of the flows are TLS encrypted. Examples of works that used this dataset include \cite{boa_conf, quicclassificationfewlabels}. The \textbf{MTA} dataset is a website (blog)~\cite{MTA} that includes many types of malware infection traffic for analysis. The website contains many types of malware, such as ransomware and exploit kits. As of 2013 to date, the blog is updated daily with relevant malware traffic, continuously adding more samples to the dataset. Using Intrusion-Detection Systems (IDS) and Antivirus software, every binary file in the PCAPs has been confirmed as malicious. Papers such as \cite{citemtatls, MTAKDD19, maldist_ccnc} have used this dataset for malware detection.

\section{Results}
In this section, we present our experimental results, which include multiple experiments on BOA and MTA datasets. For each dataset, we first find the optimal number of packets from which results do not significantly improve, then calculate the minimal features' set by performing a feature selection \cite{featuresel} (for more details on the process of the feature selection please visit Github project page\cite{featureselection}. Moreover, we present the results of our model with new samples, and the influence of the searching algorithm on the results.
Note that we have chosen statistical features because in the near future, the traffic will not include clear indicators about the service name, so we have to identify it using its statistical footprint. 

\subsection{ CBR Vs RF - BOA dataset}
\label{sec:MLComparisonMaxPer}
\begin{figure}[H]
\centering
\includegraphics[width=.8\linewidth]{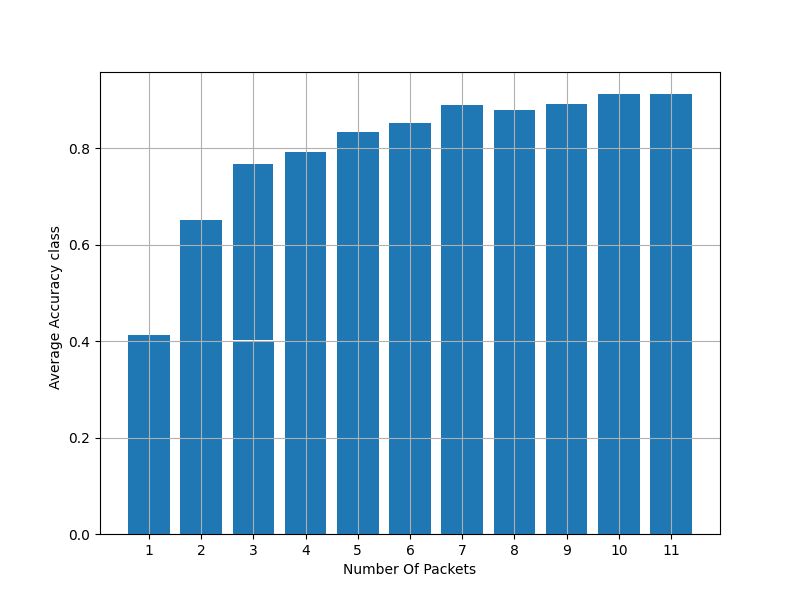}
\caption{CBR accuracy results as a function of the
number of packets - BOA dataset. As can be seen, the accuracy stops improving after 10 packets.}
\label{fig_boa_precision}
\end{figure}

In the first experiment, we wanted to check the influence of the number of packets on our solution compared to RF. Therefore, we used the entire feature sets (see Table \ref{tab_features}) and increased the number of packets from each flow until the accuracy of our solution stopped improving. 
Figure \ref{fig_boa_precision} depicts the accuracy results of our approach as a function of the number of packets. These results motivated us to use only 10 packets from each flow.
After setting the number of packets, we wanted to compare our approach to RF, as can be seen in Table \ref{tab_aknn_boa_browser_full}. The Table presents the results of our ANN model for the BOA dataset with the optimal number of processed packets (10 packets) vs RF.
\begin{table}[!]
\begin{tabular}{|p{3cm}|lll|lll|}
\hline
     {\textbf{BOA Classes}}  & \multicolumn{3}{c|}{\textbf{Our Approach}}                            & \multicolumn{3}{c|}{\textbf{RF}}                                      \\ \hline
       & \multicolumn{1}{l|}{Prec} & \multicolumn{1}{l|}{Rec}  & F1   & \multicolumn{1}{l|}{Prec} & \multicolumn{1}{l|}{Rec}  & F1   \\ \hline
win    & \multicolumn{1}{l|}{0.96} & \multicolumn{1}{l|}{0.95} & 0.95 & \multicolumn{1}{l|}{1}    & \multicolumn{1}{l|}{0.99} & 0.99 \\ \hline
OSX    & \multicolumn{1}{l|}{0.92} & \multicolumn{1}{l|}{0.97} & 0.94 & \multicolumn{1}{l|}{0.92} & \multicolumn{1}{l|}{0.99} & 0.95 \\ \hline
Linux  & \multicolumn{1}{l|}{0.98} & \multicolumn{1}{l|}{0.98} & 0.98 & \multicolumn{1}{l|}{1}    & \multicolumn{1}{l|}{0.98} & 0.99 \\ \hline
CR     & \multicolumn{1}{l|}{0.91} & \multicolumn{1}{l|}{0.90} & 0.91 & \multicolumn{1}{l|}{0.96} & \multicolumn{1}{l|}{0.98} & 0.97 \\ \hline
FF     & \multicolumn{1}{l|}{0.91} & \multicolumn{1}{l|}{0.91} & 0.91 & \multicolumn{1}{l|}{1}    & \multicolumn{1}{l|}{0.99} & 0.99 \\ \hline
IE     & \multicolumn{1}{l|}{0.97} & \multicolumn{1}{l|}{0.97} & 0.97 & \multicolumn{1}{l|}{0.97} & \multicolumn{1}{l|}{0.98} & 0.97 \\ \hline
Safari & \multicolumn{1}{l|}{0.95} & \multicolumn{1}{l|}{0.97} & 0.96 & \multicolumn{1}{l|}{1}    & \multicolumn{1}{l|}{1}    & 1    \\ \hline
\end{tabular}
 \caption{CBR model for BOA dataset using 10 packets. The results are very close to RF's results.}
 \label{tab_aknn_boa_browser_full}
\end{table}

\begin{table}[!]
\begin{tabular}{|l|p{6 cm}|}\hline
\textbf{Feature} & \textbf{Description} \\\hline
dst2src mean piat                            & Minimal arrival time between packets from the dst to src    \\\hline
ps 9                             & Packet size of the 9th packet    \\\hline
ps 21                       &   Packet size of the 21th packet     \\\hline
ps 24    &  Packet size of the 24th packet  \\\hline
ps 28                                  &     Packet's size of the 28th packet        \\\hline
beacon 10                &    The 10th index of max packet size (dst \& src)                 \\\hline
beacon 11                & The 11th index of max packet size (dst \& src)                       \\\hline
beacon 15                         &     The 15th index of max packet size (dst \& src)              \\\hline
wavelet 0                         &  Using the first coefficient on FFT of the packet sizes                  \\\hline
wavelet 16                            &   Using the 16th coefficient on FFT of the packet sizes\\\hline
wavelet 7                       &   Using the 7th coefficient on FFT of the packet sizes    \\\hline
\end{tabular}
\caption{Feature selection for BOA. These features are the minimal features' set used for BOA dataset classification for CBR classification.}
\label{tab_features_selection_boa}
\end{table}

\begin{table}[!]
    \begin{tabular}{|p{1.2cm}|l|l|l|l|l|l|l|l|}
    \hline
         & \textbf{Pr}  & \textbf{Rec.}  & \textbf{F1-SC} & \textbf{RF Pr} & \textbf{RF Rec.} & \textbf{RF F1-SC}\\ \hline
        \textbf{Win} & 0.96 & 0.95 & 0.95 & 0.99 & 0.99 & 0.99\\ \hline
        \textbf{OSX} & 0.86 & 0.9 & 0.92 & 0.92 & 0.98 & 0.95\\ \hline
        \textbf{Linux} & 0.98 & 0.98 & 0.98 & 0.99 & 0.98 & 0.99\\ \hline
        %\textbf{Accuracy} & ~ & 0.96 & ~ \\ 
        \textbf{CR} & 0.9 & 0.9 & 0.9 & 0.96 & 0.97 & 0.96\\ \hline
        \textbf{FF} & 0.91 & 0.91 & 0.91 & 0.99 & 0.98 & 0.99\\ \hline
        \textbf{IE} & 0.97 & 0.98 & 0.97 & 0.97 & 0.98 & 0.97\\ \hline
        \textbf{Safari} & 0.95 & 0.95 & 0.95 & 0.99 & 0.99 & 0.99\\ \hline
        %\textbf{Accuracy} & ~ & 0.93 & ~ \\ \hline
    \end{tabular}
        \caption{CBR model for BOA dataset using 10 packets with minimal features. The results are very close to RF's results.}
\label{tab_aknn_boa_browser}
\end{table}
\begin{table}[!]
\begin{tabular}{|l|l|l|l|l|l|l|}
\hline
         & \textbf{Pr}  & \textbf{Rec.}  & \textbf{F1-SC} & \textbf{RF Pr}  & \textbf{RF Rec.}  & \textbf{RF F1-SC} \\\hline
        \textbf{Infostealer} & 0.95 & 0.96 & 0.95 & 0.96 & 0.97 & 0.96\\\hline 
        \textbf{Dropper} & 0.93 & 0.92 & 0.92 & 0.95 & 0.95 & 0.95\\ \hline
        \textbf{ACC} & ~ & 0.96 & ~ & ~ & 0.97 & ~\\ \hline \hline
        \hline
        \textbf{Dridex} & 0.84 & 0.79 & 0.81 & 0.86 & 0.85 & 0.55\\ \hline
        \textbf{Emotet} & 0.92 & 0.93 & 0.92 & 0.93 & 0.94 & 0.93\\ \hline
        \textbf{Hancitor} & 0.96 & 0.96 & 0.96 & 0.96 & 0.96 & 0.96\\ \hline
        \textbf{Icedid} & 0.85 & 0.93 & 0.84 & 0.87 & 0.92 & 0.87\\ \hline
        \textbf{Qakbot} & 0.92 & 0.94 & 0.93 & 0.92 & 0.94 & 0.93\\
        \hline
        \textbf{Valak} & 0.92 & 0.94 & 0.93 & 0.92 & 0.94 & 0.94\\
        \hline
        \textbf{zloader} & 0.86 & 0.88 & 0.87 & 0.87 & 0.88 & 0.87\\
        \hline 
        \textbf{ACC} & ~ & 0.93 & ~ & ~ & 0.95 & ~\\
        \hline
    \end{tabular}
     \caption{CBR model for MTA dataset using 98 packets}
    \label{tab_aknn_mta}
\end{table}
\begin{table}[!]
\begin{tabular}{|l|l|l|l|l|l|l|}
\hline
 & \textbf{Pr} & \textbf{Rec.} & \textbf{F1-SC} & \textbf{RF Pr} & \textbf{RF Rec.} & \textbf{RF F1-SC}\\ \hline
\textbf{Infostealer} & 0.95 & 0.96 & 0.95 & 0.95 & 0.97 & 0.95\\ \hline
\textbf{Dropper} & 0.93 & 0.92 & 0.92 & 0.94 & 0.94 & 0.94\\ \hline
\textbf{Accuracy} & \multicolumn{6}{c|}{0.96} \\ \hline \hline
\textbf{Dridex} & 0.84 & 0.79 & 0.81 & 0.85 & 0.82 & 0.82\\ \hline
\textbf{Emotet} & 0.92 & 0.93 & 0.93 & 0.93 & 0.93 & 0.93\\ \hline
\textbf{Hancitor} & 0.95 & 0.95 & 0.95 & 0.95 & 0.95 & 0.95 \\ \hline
\textbf{Icedid} & 0.85 & 0.92 & 0.84 & 0.86 & 0.92 & 0.86\\ \hline
\textbf{Qakbot} & 0.92 & 0.93 & 0.93 & 0.93 & 0.93 & 0.93 \\ \hline
\textbf{Valak} & 0.92 & 0.93 & 0.93 & 0.92 & 0.93 & 0.93\\ \hline
\textbf{zloader} & 0.86 & 0.88 & 0.87 & 0.87 & 0.88 & 0.87 \\ \hline
\textbf{Accuracy} & \multicolumn{6}{c|}{0.93} \\ \hline
\end{tabular}
\caption{CBR model minimal features set for the MTA dataset using 98 packets}
\label{tab_aknn_mta_feature_selection}
\end{table}
\begin{table}[!]
    \begin{tabular}{|l|l|l|l|l|}
    \hline
         & \textbf{F1-Score  (before adding new class)}  &  \textbf{F1-Score} \\\hline
        \textbf{Beacon} &  & 0.95 \\ \hline
        \textbf{Infostealer} & 0.97 & 0.96 \\\hline
        \textbf{Dropper} & 0.92 & 0.91 \\\hline \hline
        \textbf{Cobalt Strike} &  & 0.95 \\\hline
        \textbf{Dridex} & 0.81 & 0.79 \\ \hline
        \textbf{Emotet} & 0.93 & 0.93 \\ \hline
        \textbf{Hancitor} &  0.97 & 0.97 \\ \hline
        \textbf{Icedid} & 0.84 & 0.82 \\ \hline
        \textbf{Qakbot} & 0.93& 0.93 \\ \hline
        \textbf{Valak} & 0.93& 0.93 \\ \hline
        \textbf{zloader} & 0.87& 0.84 \\ \hline
        
    \end{tabular}
    \caption{CBR model for MTA's results for Cobalt Strike. The results show, that accuracy of identifying current classes is not worsen, by the addition of the new class.} 
    \label{tab_aknn_cobalt_strike}
\end{table}
Note that we do not aim to beat RF, but rather perform as close as possible to RF while our solution does not need retraining once a new class appears. Our results confirm this by showing that our CBR classification results are very close to the results that RF provides, which was one of our goals in this experiment. 

\subsection{CBR Vs RF - Minimal Features set on BOA dataset}
\label{sec:MLComparisonMaxPerMinF}
In the second evaluation, we did a supervised filter-based feature selection \cite{featuresel}, to obtain the minimal number of features that provide maximal performance using the 10 packets from each sample. The results of both algorithms (ours and RF) with the minimal feature set used for both classifiers are depicted in Table \ref{tab_aknn_boa_browser}, while the feature set is depicted in Table \ref{tab_features_selection_boa}, and as we can see that RF performs slightly better than our algorithm. From the above experiments, we see that based on the BOA dataset, our CBR model achieved results that are close to RF. 
\subsection{CBR with OOD Vs RF - BOA dataset}
\label{sec:MLOODBOAApplications}
\begin{figure}[H]
    \centering
    \includegraphics[height=3cm,width=9cm]{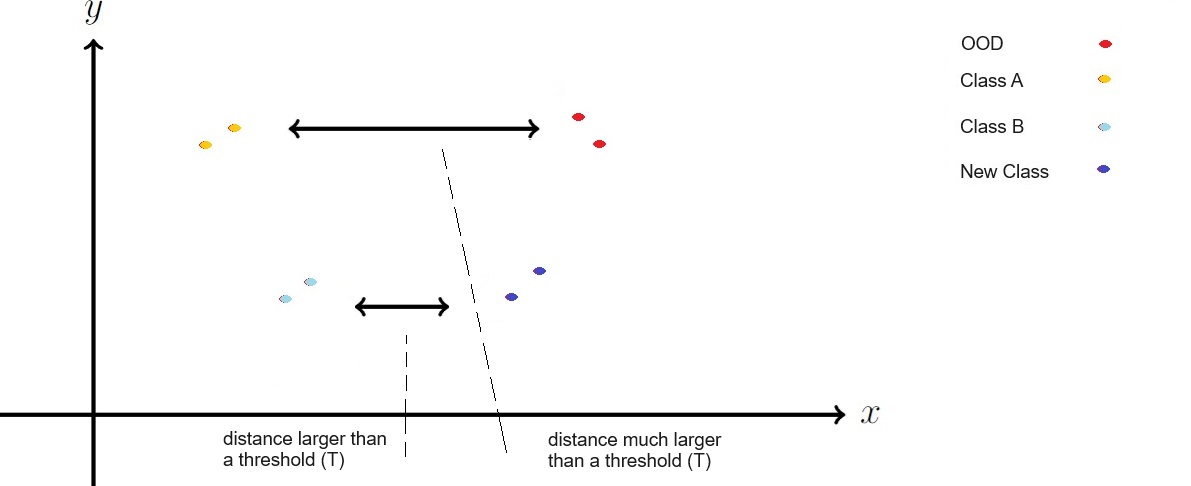}
    \caption{CBR with OOD. Classify samples as new classes or OODs according to their distances from the nearest classes.}
    \label{fig:CBRWITHOODFLOW}
\end{figure}

In this section, we have used a different strategy. We used labeled applications, and their features, and used all their features as is, instead of calculating features from packet capture files, while some of these applications are labeled as not categorized (e.g. Netflix traffic), as shown in Table \ref{fig_acbr_methodology}. To complete the picture, note that each sample contains its features, and each one contains packet size, and flow direction (inbound/outbound). Afterward, we used CBR and labeled the samples whose distances from their class are larger than a threshold as OOD \cite{jeong2020ood} as depicted in Figure 
\ref{fig:CBRWITHOODFLOW} , and the results are depicted in Table \ref{tab_boa_applications}: 
\begin{table}[H]
    \begin{tabular}{|p{1.4cm}|l|l|l|l|l|l|l|l|}
    \hline
        \textbf{Class}& Pr  & Rec.  & F1-SC & RF Pr  & RF Rec.  & RF F1-SC\\\hline
        \textbf{Dropbox} & 0.9 & 1.0 & 0.95 & 0.91 & 0.91 & 0.91\\ 
        \textbf{Facebook} & 0.97 & 1.0 & 0.98 & 0.675 & 0.675 & 0.675\\ 
        \textbf{google} & 0.79 & 1.0 & 0.88 & 0.95 & 0.95 & 0.95 \\ 
        \textbf{Microsoft} & 1.0 & 1.0 & 1.0& 0.91 & 0.91 & 0.91\\ 
        \textbf{TeamViewer} & 0.99 & 1.0 & 1.0  & 0.95 & 0.95 & 0.95\\ 
        \textbf{Twitter} & 0.99 & 0.98 & 0.98 & 0.99 & 0.99 & 0.99\\ 
        \textbf{Youtube} & 0.67 & 0.88 & 0.76  & 0.91 & 0.91 & 0.91\\ \hline
    \end{tabular}
    \caption{BOA applications CBR classification. As seen in the table, in some cases CBR provides more accurate results than RF, and in some cases, it is vice versa.}
\label{tab_boa_applications}
\end{table}
\iffalse
\begin{figure}[H]
    \centering
    \includegraphics[width=.8\linewidth]{images/oodflow.jpg}
    \caption{Caption}
    \label{fig:OODFLOW}
\end{figure}
\begin{table}[H]
    \begin{tabular}{|p{1.2cm}|l|l|}
    \hline
        \textbf{Class}& Features\\\hline
        \textbf{dropbox} & [0, 0], [1, 0], [0, 131], [1, 0], [1, 1360], [1, 1360],  [0, 0] \\ 
        \textbf{} & [1, 60], [0, 0], [0, 134], [1, 59], [0, 0], [0, 362], [0, 714]\\
        \textbf{} & [0, 0], [0, 0], [0, 0], [0, 0], [0, 0], [0, 0], [0, 0], [0, 0], [0, 0]\\
        \textbf{} & [0, 0], [0, 0], [0, 0], [0, 0], [0, 0], [0, 0], [0, 0], [0, 0], [0, 0]\\
        \textbf{} & [0, 0], [0, 0], [0, 0], [0, 0], [0, 0], [0, 0], [0, 0], [0, 0], [0, 0]\\
        \textbf{} & [0, 0], [0, 0], [0, 0], [0, 0], [0, 0], [0, 0], [0, 0], [0, 0], [0, 0]\\
        \textbf{} & [0, 0], [0, 0], [0, 0], [0, 0], [0, 0], [0, 0], [0, 0], [0, 0], [0, 0]\\
        \textbf{} & [0, 0], [0, 0], [0, 0], [0, 0], [0, 0], [0, 0], [0, 0], [0, 0], [0, 0]\\
        \textbf{} & [0, 0], [0, 0], [0, 0], [0, 0], [0, 0], [0, 0], [0, 0], [0, 0], [0, 0]\\
        \textbf{} & [0, 0], [0, 0], [0, 0], [0, 0], [0, 0], [0, 0], [0, 0], [0, 0]\\ 
        \textbf{} & [0, 0], [0, 0], [0, 0], [0, 0], [0, 0], [0, 0], [0, 0], [0, 0]\\ 
        \textbf{} & [1, 1360], [0, 0], [0, 0], [0, 0], [0, 0], [0, 0], [0, 0] \\ \hline
    \end{tabular}
    \caption{Sample application's test with related features}
\label{tab_application_sample}
\end{table}
\fi

As we can see from the results, for some classes our algorithm outperforms RF, and for others it is vice versa.

\subsection{CBR Vs RF - MTA dataset}
\label{sec:MLComparisonMaxPerMTA}
\begin{figure}[H]
\centering
\includegraphics[width=.8\linewidth]{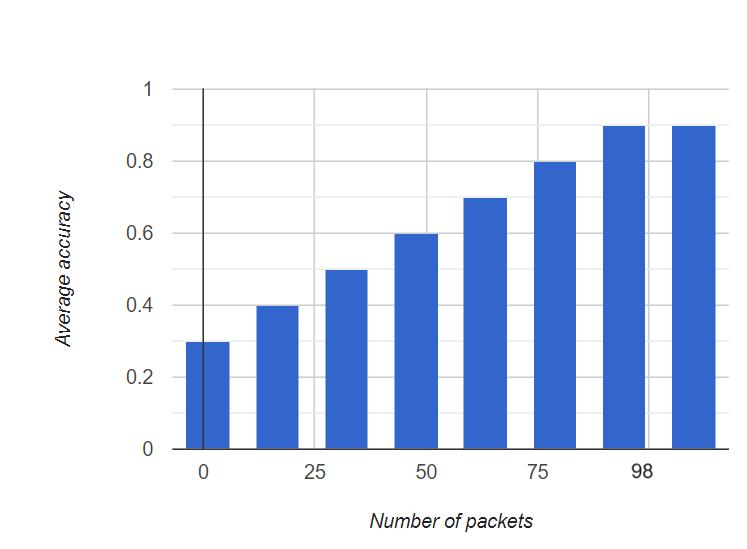}
\caption{CBR accuracy results as a function of the
number of packets - MTA dataset. Notice, that after 98 packets, the accuracy does not improve.}
\label{fig:ACBRMTAPlotBar}
\end{figure}

In this section, similar to the first experiment, we wanted to check the influence of the number of packets on our solution. Therefore, we used the entire feature sets and increased the number of packets from each flow until the accuracy of our solution stopped improving (this happened after 98 packets), as depicted in Figure \ref{fig:ACBRMTAPlotBar}. We present the optimal classifier results, using 98 packets (on average around 50-60 KB) for MTA dataset in Table \ref{tab_aknn_mta}. Again, we see that RF as an ensemble method slightly outperforms our CBR. Notice, that MTA consists of a main class, which includes InfoStealer and Dropper, and the rest are secondary classes. 
\subsection{\label{sec:MLComparisonMaxPerMinFMTA} CBR Vs RF - Minimal Features set on MTA dataset}
In this section, we performed a supervised filter-based feature selection \cite{featuresel} on MTA dataset, to obtain the minimal number of features, which provide maximal performance using 98 packets. The results are depicted in Table \ref{tab_aknn_mta_feature_selection} (the list of features are the same features for RF and CBR, and are depicted in Table \ref{tab_features_selection_mta_browser}).

\begin{table*}[ht]
\begin{tabular}{|l|l|l|}\hline
\textbf{Feature} & \textbf{Description} \\\hline
window delta 11                             & In a 5 seconds window, the 11th index of the max difference between packet size sent/received in the same direction    \\\hline
wavelet 9                       &   Using the 9th coefficient on FFT of the packet sizes in a 10 seconds window     \\\hline
ps 29    &  Packet size of the 29th packet our of first 30 packets in the same direction \\\hline
ps 6                                  &     Packet's size of the 6th packet our of first 30 packets in the same direction        \\\hline
wavelet 12                &    Using the 12th coefficient on FFT of the packet sizes in a 10 seconds window                  \\\hline
window delta 10                & In a 5 seconds window, the 10th index of the max difference between packet's size sent/received in the same direction                        \\\hline
wavelet 18                         &     Using the 18th coefficient on FFT of the packet sizes in a 10 seconds window              \\\hline
bidirectional mean piat ms                         & average arrival time between packets in both directions                   \\\hline
wavelet 5                            &  Using the 5th coefficient on FFT of the packet sizes in a 10 seconds window \\\hline
ps 4                       &  Packet's size of the 4th packet our of first 30 packets in the same direction     \\\hline
\end{tabular}
\caption{Feature selection for MTA. These features' are
the minimal features set used for MTA dataset
for CBR classification. Note, that the number near each features shows the granularity of its usage, for example, \textbf{ps 4}, means the 4th packet's size of the first 30 packets in the same direction. }
\label{tab_features_selection_mta_browser}
\end{table*}

From the above experiments,  we can see that also based on the MTA dataset, our CBR model achieved results that are close to RF. Furthermore, after presenting that our model results are close to RF, in the following section we will show the ability of our model to adapt to new classes without any retraining phase.
\subsection{\label{sec:MLComparisonACBR} CBR New Classes Classification}
So far we tested our approach on known classes. In the following section, we present the results of our approach in the case of new classes of malware (i.e. Cobalt Strike), first seen by the model. In this experiment, we did not mention RF results, since it is not able to deal with new classes dynamically, only predict based on trained classes. The result are shown in Table \ref{tab_aknn_cobalt_strike}, from the results, we can see there is a slight decrease in the accuracy when testing samples from new classes, however, all the classes are classified accurately. 

\subsection{\label{sec:ANNBenchmark}ANN Benchmark}
\begin{figure*}[!]
    \centering
    \begin{subfigure}[t]{0.45\textwidth}
        \centering
        \includegraphics[width=9cm, height=6cm]{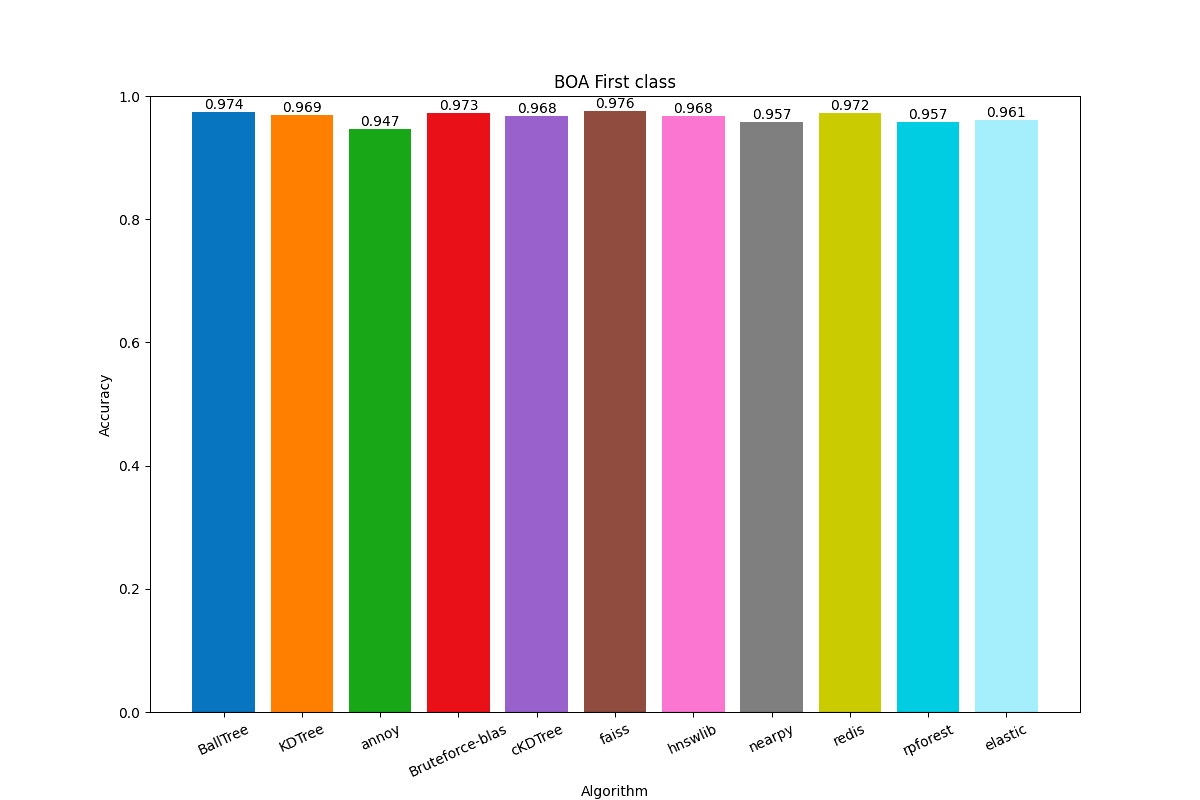}
        \caption{BOA First Class's ANN Benchmark}
        \label{fig_boa_benchmark}
    \end{subfigure}
    \begin{subfigure}[t]{0.45\textwidth}
        \centering
        \includegraphics[width=9cm, height=6cm]{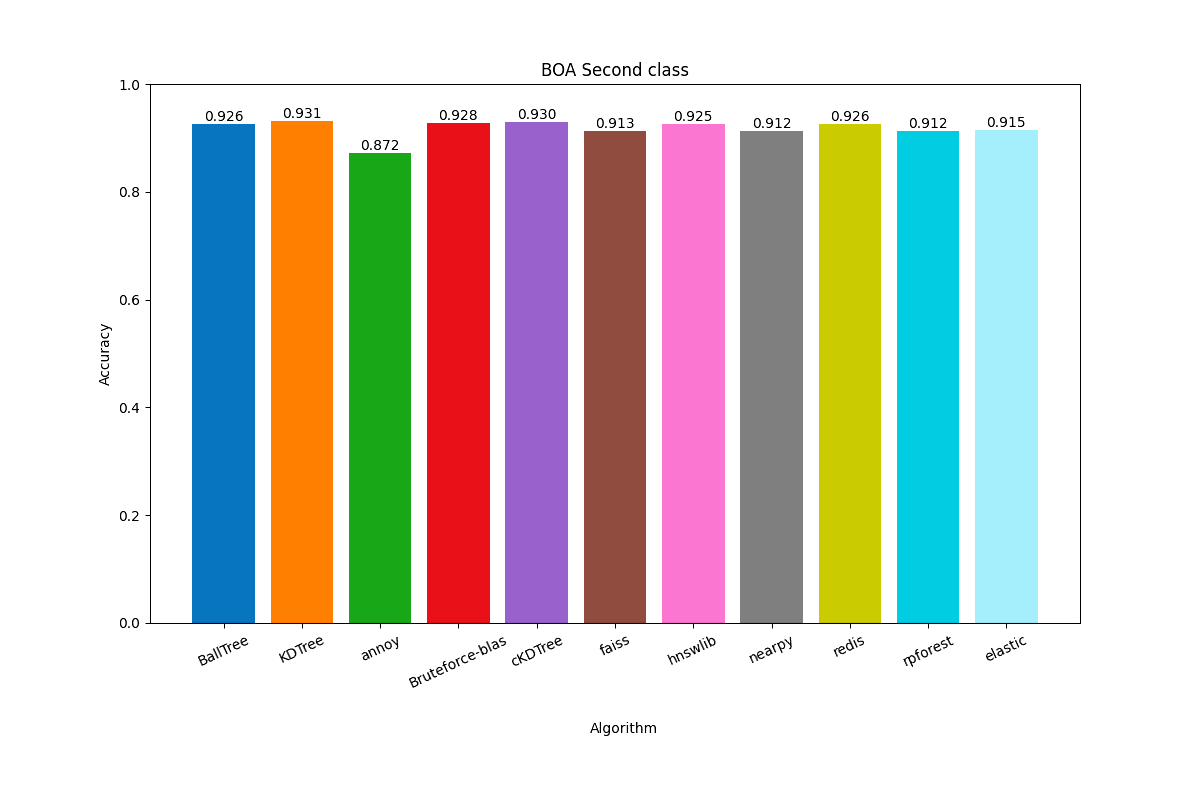}
        \caption{BOA Second Class's ANN Benchmark}
        \label{fig_boa_benchmark_2nd_class}
    \end{subfigure}
   
    \begin{subfigure}[t]{0.45\textwidth}
        \centering
        \includegraphics[width=9cm, height=6cm]{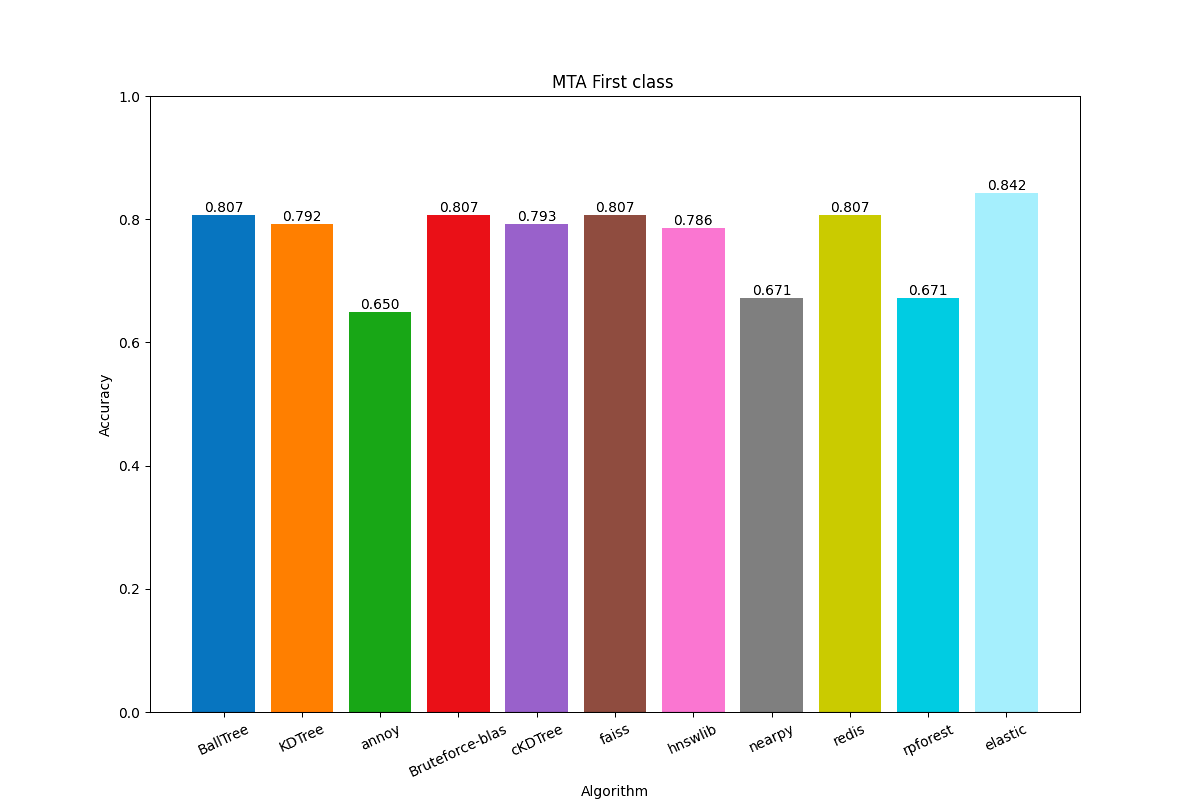}
        \caption{MTA First Class's ANN Benchmark}
        \label{fig_mta_benchmark}
    \end{subfigure}
   ~ 
    \begin{subfigure}[t]{0.45\textwidth}
        \centering
        \includegraphics[width=9cm, height=6cm]{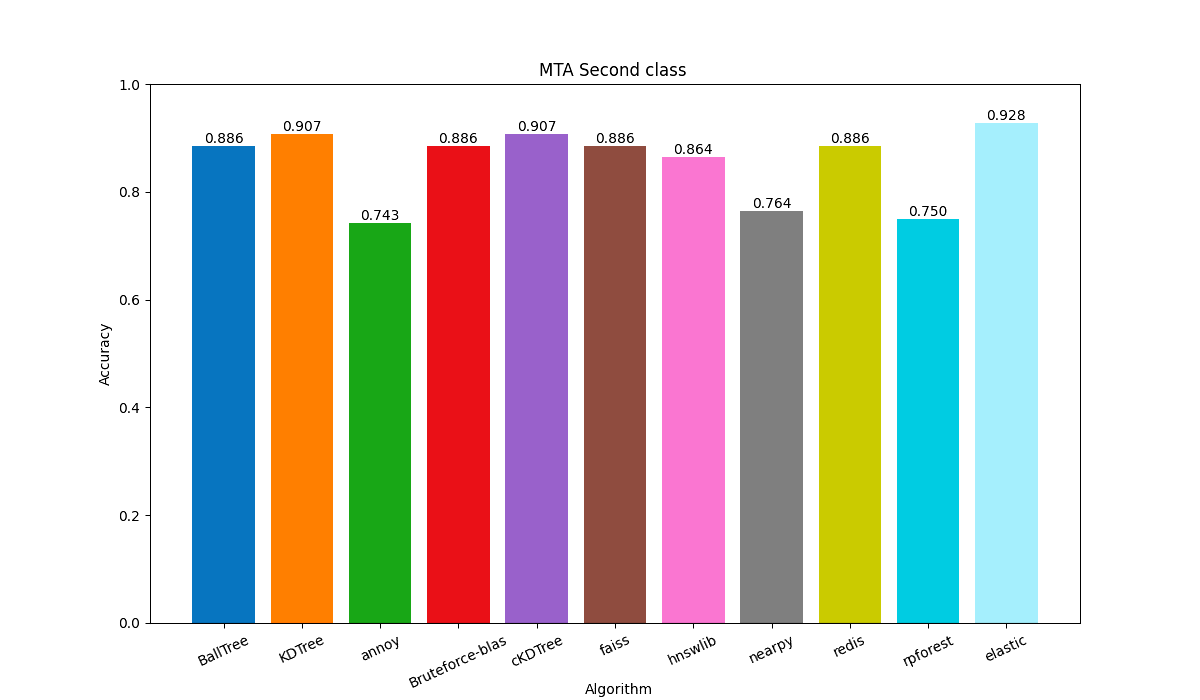}
        \caption{MTA Second Class's ANN Benchmark}
        \label{fig_mta_benchmark_2nd_class}
    \end{subfigure}
    \caption{ANN-Benchmarks. Notice, that for the MTA dataset, elastic, which is the search algorithm we use provides the best results. As for the   BOA dataset, for which brute force approach provides the best results. This can be the case, in which classes are close to each other, so a brute force approach can work well.}
    \label{fig_ann_benchmark}
\end{figure*}

In this section, we provide our final evaluation which is a comparison between the ANN search algorithms. We have used vector search, which provided the best results in most of the experiments we did, however, in some cases, the best results were produced by the Ball tree \cite{balltree} algorithm. The commonly used ANN algorithms are shown in Figure~\ref{fig_ann_benchmark}.
The list of search algorithms is as follows:
\begin{enumerate}
    \item Ball tree - partitions data points into a nested set of balls (a ball is a solid figure bounded by a sphere).
    \item KD tree - a binary tree, in which each node represents a k's dimensions point \cite{mount1998ann}.
    \item annoy - uses file-based data structures mapped into the memory, to allow searching for points, which are close to a query point \cite{annoy2018}.
    \item Bruteforce - blas - Uses brute force to calculate Euclidean distances and sorts the results \cite{garcia2010k}.
    \item cKDTree - improved kd tree construction algorithm \cite{ramasubramani2020freud}.
    \item faiss - AI similarity search \cite{chen2019robustiq}.
    \item hnswlib - fast approximate nearest neighbor search \cite{rahman2022evaluating}.
    \item nearpy - a simple yet modular little framework for ANN search \cite{nearpy2013}.
    \item redis - uses vector similarity in a key/value in memory database \cite{zhang2019sextant}.
    \item rpforest - uses a forest of random projection trees \cite{yan2019k}.
    \item elastic - incorporates KD trees to support searches on geospatial and numeric data.
\end{enumerate}
From the results, we can see that the algorithm we have used gives the best results for the MTA dataset, or is close to the best possible results, in the case of the BOA dataset, in which Bruteforce - blas gave the best results. We chose elastic, since it not only usually provides the best results  for the selected datasets, but also has a near real-time speed for processing samples.

\section{\label{sec:Conclusions}Discussions and Future Perspectives}
%Encrypted traffic classification is an invaluable part of cybersecurity, since network traffic encryption has become prevalent. Recently, there has been a massive change in the internet protocols, where new network protocols such as QUIC \cite{QUIC}, HTTP/3, and new privacy-concerned protocols such as TLS 1.3 and DoH \cite{DoH} have been introduced.
%Retraining ML (and especially DL) models every time they become obsolete is both resource and time-consuming, especially when the application is complex, and the evaluated datasets are large. Therefore, there is a growing need to allow classification models to detect and adapt to new classes dynamically, without retraining.

In this paper, we have shown how our approach, which relies on the ANN method, can achieve the following goals - detect malware activity on encrypted network traffic, classify the malware type and name, create a simple model that can be maintained easily, detect unknown traffic, classify it as new classes, creates an alternative solution for retraining.
Our plans for future work include: features selection for behavioral features needed for malware classification, features selection for behavioral features needed for application classification, feature selection for behavioral features needed for application type classification (chat, VoIP \cite{voip}, data download, VOD \cite{vod}, etc.), 
understanding the impact of the order of the features, on the classification's results, 
optimize runtime performance, use \textit{CBR}'s classification on additional datasets, benchmark our solution VS additional ML/DL methods, as we can see in Figure \ref{fig_ann_benchmark}, compare to additional ANN search algorithms, our algorithm provided the best results most of the time. In addition, CBR also uses the elastic database, which provides very fast data access, and fast run-time performance. Furthermore, To the best of our knowledge, this is the first paper to show usage of few shots learning, and OOD, and this can be a basis for future research, in order to improve few shots learning accuracy.

\iffalse
\section{Steps}
\begin{enumerate}
\item Choosing behavioral features - classified until layer 3 and above layer 3 (and marking TLS features).
\item Creating dataset.
\item  Integrating flows into several search engines by using approximate nearest neighbor (ANN).
\item Benchmarking the search engines.
\item Compare results with goals
\end{enumerate}
\fi

\section{Acknowledgment}
 This work was supported by the Ariel Cyber Innovation Center in conjunction with the Israel National Cyber Directorate in the Prime Minister's Office, the Israel Innovation Authority, and the ENTM consortium, and is patent-based on US Provisional Patent Application No. 63/448,460. 
\bibliographystyle{IEEEtran}  
\bibliography{main}

\begin{thebibliography}{10}
\providecommand{\url}[1]{#1}
\csname url@samestyle\endcsname
\providecommand{\newblock}{\relax}
\providecommand{\bibinfo}[2]{#2}
\providecommand{\BIBentrySTDinterwordspacing}{\spaceskip=0pt\relax}
\providecommand{\BIBentryALTinterwordstretchfactor}{4}
\providecommand{\BIBentryALTinterwordspacing}{\spaceskip=\fontdimen2\font plus
\BIBentryALTinterwordstretchfactor\fontdimen3\font minus
  \fontdimen4\font\relax}
\providecommand{\BIBforeignlanguage}[2]{{%
\expandafter\ifx\csname l@#1\endcsname\relax
\typeout{** WARNING: IEEEtran.bst: No hyphenation pattern has been}%
\typeout{** loaded for the language `#1'. Using the pattern for}%
\typeout{** the default language instead.}%
\else
\language=\csname l@#1\endcsname
\fi
#2}}
\providecommand{\BIBdecl}{\relax}
\BIBdecl

\bibitem{wang2020generalizing}
Y.~Wang, Q.~Yao, J.~T. Kwok, and L.~M. Ni, ``Generalizing from a few examples:
  A survey on few-shot learning,'' \emph{ACM computing surveys (csur)},
  vol.~53, no.~3, pp. 1--34, 2020.

\bibitem{nlp}
\BIBentryALTinterwordspacing
Oracle, ``Quic transport protocol rfc9000,'' 2023. [Online]. Available:
  \url{https://www.oracle.com/il-en/artificial-intelligence/what-is-natural-language-processing/}
\BIBentrySTDinterwordspacing

\bibitem{goodman2020packet2vec}
E.~L. Goodman, C.~Zimmerman, and C.~Hudson, ``Packet2vec: Utilizing word2vec
  for feature extraction in packet data,'' \emph{arXiv preprint
  arXiv:2004.14477}, 2020.

\bibitem{li2020weighted}
J.~Li, H.~Zhang, and Z.~Wei, ``The weighted word2vec paragraph vectors for
  anomaly detection over http traffic,'' \emph{IEEE Access}, vol.~8, pp.
  141\,787--141\,798, 2020.

\bibitem{corizzo2020feature}
R.~Corizzo, E.~Zdravevski, M.~Russell, A.~Vagliano, and N.~Japkowicz, ``Feature
  extraction based on word embedding models for intrusion detection in network
  traffic,'' \emph{Journal of Surveillance, Security and Safety}, vol.~1,
  no.~2, pp. 140--150, 2020.

\bibitem{bar2022simcse}
R.~Bar and C.~Hajaj, ``Simcse for encrypted traffic detection and zero-day
  attack detection,'' \emph{IEEE Access}, 2022.

\bibitem{flowpic2021}
T.~Shapira and Y.~Shavitt, ``Flowpic: {A} generic representation for encrypted
  traffic classification and applications identification,'' \emph{{IEEE} Trans.
  Netw. Serv. Manag.}, vol.~18, no.~2, pp. 1218--1232, 2021.

\bibitem{deepmal}
G.~Mar{\'{\i}}n, P.~Casas, and G.~Capdehourat, ``Deepmal - deep learning models
  for malware traffic detection and classification,'' \emph{CoRR}, vol.
  abs/2003.04079, 2020.

\bibitem{horowicz2022few}
E.~Horowicz, T.~Shapira, and Y.~Shavitt, ``A few shots traffic classification
  with mini-flowpic augmentations,'' in \emph{IMC}, 2022, pp. 647--654.

\bibitem{quic}
\BIBentryALTinterwordspacing
Google, ``Quic, a multiplexed transport over udp,'' 2023. [Online]. Available:
  \url{https://www.chromium.org/quic/}
\BIBentrySTDinterwordspacing

\bibitem{bottger2019empirical}
T.~B{\"o}ttger, F.~Cuadrado, G.~Antichi, E.~L. Fernandes, G.~Tyson, I.~Castro,
  and S.~Uhlig, ``An empirical study of the cost of dns-over-https,'' in
  \emph{Proceedings of the Internet Measurement Conference}, 2019, pp. 15--21.

\bibitem{flowclassification}
\BIBentryALTinterwordspacing
D.~Serpanos and T.~Wolf, ``Transport layer systems,'' 2011. [Online].
  Available:
  \url{https://www.sciencedirect.com/topics/computer-science/flow-classification}
\BIBentrySTDinterwordspacing

\bibitem{WeiWangMalwareTrafficClassification}
W.~Wang, M.~Zhu, X.~Zeng, X.~Ye, and Y.~Sheng, ``Malware traffic classification
  using convolutional neural network for representation learning,'' in
  \emph{{ICOIN}}, 2017, pp. 712--717.

\bibitem{aws_update_neural_network}
\BIBentryALTinterwordspacing
A.~Moschitti. Updating neural networks to recognize new categories, with
  minimal retraining. [Online]. Available:
  \url{https://www.amazon.science/blog/updating-neural-networks-to-recognize-new-categories-with-minimal-retraining}
\BIBentrySTDinterwordspacing

\bibitem{schulz2020extending}
J.~Schulz, C.~Veal, A.~Buck, D.~Anderson, J.~Keller, M.~Popescu, G.~Scott,
  D.~K. Ho, and T.~Wilkin, ``Extending deep learning to new classes without
  retraining,'' in \emph{DSMEOBT XXV}, vol. 11418, 2020, pp. 13--26.

\bibitem{korn1998fast}
F.~Korn, N.~Sidiropoulos, C.~Faloutsos, E.~Siegel, and Z.~Protopapas, ``Fast
  nearest neighbor search in medical image databases,'' 1998.

\bibitem{bustos2006graphics}
B.~Bustos, O.~Deussen, S.~Hiller, and D.~Keim, ``A graphics hardware
  accelerated algorithm for nearest neighbor search,'' in \emph{ICCS}, 2006,
  pp. 196--199.

\bibitem{beatenpath}
L.~Boytsov, D.~Novak, Y.~Malkov, and E.~Nyberg, ``Off the beaten path: Let's
  replace term-based retrieval with k-nn search,'' in \emph{CIKM}, New York,
  NY, USA, 2016, p. 1099–1108.

\bibitem{aumuller2020ann}
M.~Aum{\"u}ller, E.~Bernhardsson, and A.~Faithfull, ``Ann-benchmarks: A
  benchmarking tool for approximate nearest neighbor algorithms,''
  \emph{Information Systems}, vol.~87, p. 101374, 2020.

\bibitem{ANN_ALGORITHM}
Apache, ``Approximate nearest neighbors algorithm,'' 2023.

\bibitem{distmetric}
\BIBentryALTinterwordspacing
S.~A. Gokte, ``Most popular distance metrics used in knn and when to use
  them,'' 2023. [Online]. Available:
  \url{https://www.kdnuggets.com/2020/11/most-popular-distance-metrics-knn.html}
\BIBentrySTDinterwordspacing

\bibitem{weiwangmalwaretrafficclassificationm2cnn2017}
W.~Wang, M.~Zhu, X.~Zeng, X.~Ye, and Y.~Sheng, ``Malware traffic classification
  using convolutional neural network for representation learning,'' in
  \emph{ICOIN}, 2017, pp. 712--717.

\bibitem{wang1dcnn}
W.~Wang, M.~Zhu, J.~Wang, X.~Zeng, and Z.~Yang, ``End-to-end encrypted traffic
  classification with one-dimensional convolution neural networks,'' in
  \emph{{ISI}, Beijing, China, July 22-24}, 2017, pp. 43--48.

\bibitem{citemtatls}
D.~Kim, J.~Han, J.~Lee, H.~Roh, and W.~Lee, ``Poster: Feasibility of malware
  traffic analysis through tls-encrypted flow visualization,'' in \emph{{ICNP}
  2020, Madrid, Spain, October 13-16}, 2020.

\bibitem{cite5}
C.~Liu, L.~He, G.~Xiong, Z.~Cao, and Z.~Li, ``Fs-net: {A} flow sequence network
  for encrypted traffic classification,'' in \emph{{INFOCOM} Paris, France,
  April 29 - May 2}, 2019, pp. 1171--1179.

\bibitem{boa_conf}
R.~Dubin, A.~Dvir, O.~Pele, J.~Muehlstein, Y.~Zion, M.~Bahumi, and
  I.~Kirshenboim, ``Analyzing https encrypted traffic to identify user’s
  operating system, browser and application,'' in \emph{CCNC}, Jun. 2017.

\bibitem{cite6}
P.~Wang, X.~Chen, F.~Ye, and Z.~Sun, ``A survey of techniques for mobile
  service encrypted traffic classification using deep learning,'' \emph{{IEEE}
  Access}, vol.~7, pp. 54\,024--54\,033, 2019.

\bibitem{conti_new_2017}
V.~F. Taylor, R.~Spolaor, M.~Conti, and I.~Martinovic, ``Robust smartphone app
  identification via encrypted network traffic analysis,'' \emph{IEEE TIFS},
  vol.~13, no.~1, pp. 63--78, Jan 2018.

\bibitem{PINHEIRO20198}
\BIBentryALTinterwordspacing
A.~J. Pinheiro, J.~{de M. Bezerra}, C.~A. Burgardt, and D.~R. Campelo,
  ``Identifying iot devices and events based on packet length from encrypted
  traffic,'' \emph{Computer Communications}, vol. 144, pp. 8--17, 2019.
  [Online]. Available:
  \url{https://www.sciencedirect.com/science/article/pii/S0140366419300052}
\BIBentrySTDinterwordspacing

\bibitem{cite3}
O.~Salman, I.~H. Elhajj, A.~I. Kayssi, and A.~Chehab, ``Data representation for
  {CNN} based internet traffic classification: a comparative study,''
  \emph{Multim. Tools Appl.}, vol.~80, no.~11, pp. 16\,951--16\,977, 2021.

\bibitem{cite10}
S.~Rezaei and X.~Liu, ``Deep learning for encrypted traffic classification: An
  overview,'' \emph{{IEEE} Commun. Mag.}, vol.~57, no.~5, pp. 76--81, 2019.

\bibitem{CGNN}
\BIBentryALTinterwordspacing
B.~Pang, Y.~Fu, S.~Ren, Y.~Wang, Q.~Liao, and Y.~Jia, ``Cgnn: Traffic
  classification with graph neural network,'' 2021. [Online]. Available:
  \url{https://arxiv.org/abs/2110.09726}
\BIBentrySTDinterwordspacing

\bibitem{MTAKDD19}
I.~Letteri, G.~D. Penna, L.~D. Vita, and M.~T. Grifa, ``Mta-kdd'19: {A} dataset
  for malware traffic detection,'' in \emph{{CEUR}, Italy, February 4-7},
  M.~Loreti and L.~Spalazzi, Eds., vol. 2597, 2020, pp. 153--165.

\bibitem{yesml2017}
A.~Demontis, M.~Melis, B.~Biggio, D.~Maiorca, D.~Arp, K.~Rieck, I.~Corona,
  G.~Giacinto, and F.~Roli, ``Yes, machine learning can be more secure! a case
  study on android malware detection,'' \emph{IEEE Transactions on Dependable
  and Secure Computing}, 2017.

\bibitem{cite30}
J.~G. de~la Puerta, I.~Pastor{-}L{\'{o}}pez, B.~Sanz, and P.~G. Bringas,
  ``Network traffic analysis for android malware detection,'' in \emph{{HAIS}},
  ser. Lecture Notes in Computer Science, vol. 11734, 2019, pp. 468--479.

\bibitem{D201950}
A.~D., V.~K. K.A., S.~C. S., and V.~P., ``Malware traffic classification using
  principal component analysis and artificial neural network for extreme
  surveillance,'' \emph{Computer Communications}, vol. 147, pp. 50--57, 2019.

\bibitem{1505198}
A.~Chan and N.~Vasconcelos, ``Classification and retrieval of traffic video
  using auto-regressive stochastic processes,'' in \emph{IEEE Proceedings.
  Intelligent Vehicles Symposium, 2005.}, 2005, pp. 771--776.

\bibitem{10.1117/12.325703}
\BIBentryALTinterwordspacing
T.~Zhang and C.-C.~J. Kuo, ``{Content-based classification and retrieval of
  audio},'' in \emph{Advanced Signal Processing Algorithms, Architectures, and
  Implementations VIII}, F.~T. Luk, Ed., vol. 3461, International Society for
  Optics and Photonics.\hskip 1em plus 0.5em minus 0.4em\relax SPIE, 1998, pp.
  432 -- 443. [Online]. Available: \url{https://doi.org/10.1117/12.325703}
\BIBentrySTDinterwordspacing

\bibitem{LICHY2023103000}
A.~Lichy, O.~Bader, R.~Dubin, A.~Dvir, and C.~Hajaj, ``When a rf beats a cnn
  and gru, together—a comparison of deep learning and classical machine
  learning approaches for encrypted malware traffic classification,''
  \emph{Computers \& Security}, vol. 124, p. 103000, 2023.

\bibitem{RF}
\BIBentryALTinterwordspacing
IBM, ``What is random forest?'' 2023. [Online]. Available:
  \url{https://www.ibm.com/topics/random-forest}
\BIBentrySTDinterwordspacing

\bibitem{chen2019all}
T.~Chen, ``All versus one: an empirical comparison on retrained and incremental
  machine learning for modeling performance of adaptable software,'' in
  \emph{SEAMS}, 2019, pp. 157--168.

\bibitem{QIBLOCKCHAIN}
H.~Qi, J.~Wang, W.~Li, Y.~Wang, and T.~Qiu, ``A blockchain-driven iiot traffic
  classification service for edge computing,'' \emph{IEEE Internet of Things
  Journal}, vol.~8, no.~4, pp. 2124--2134, 2021.

\bibitem{KNN_ALGORITHM}
IBM, ``K-nearest neighbors algorithm,'' 2023.

\bibitem{improved_knn2020}
C.~Ma, X.~Du, and L.~Cao, ``Improved knn algorithm for fine-grained
  classification of encrypted network flow,'' \emph{Electronics}, vol.~9,
  no.~2, 2020.

\bibitem{kramer2013dimensionality}
O.~Kramer, \emph{Dimensionality reduction with unsupervised nearest
  neighbors}.\hskip 1em plus 0.5em minus 0.4em\relax Springer, 2013, vol.~51.

\bibitem{huang2015query}
Q.~Huang, J.~Feng, Y.~Zhang, Q.~Fang, and W.~Ng, ``Query-aware
  locality-sensitive hashing for approximate nearest neighbor search,''
  \emph{VLDBE}, vol.~9, no.~1, pp. 1--12, 2015.

\bibitem{elastic_search}
\BIBentryALTinterwordspacing
elastic, ``What is elastic search?'' 2023. [Online]. Available:
  \url{https://www.elastic.co/what-is/elasticsearch}
\BIBentrySTDinterwordspacing

\bibitem{pmlr-v162-sun22d}
\BIBentryALTinterwordspacing
Y.~Sun, Y.~Ming, X.~Zhu, and Y.~Li, ``Out-of-distribution detection with deep
  nearest neighbors,'' in \emph{Proceedings of the 39th International
  Conference on Machine Learning}, ser. Proceedings of Machine Learning
  Research, K.~Chaudhuri, S.~Jegelka, L.~Song, C.~Szepesvari, G.~Niu, and
  S.~Sabato, Eds., vol. 162.\hskip 1em plus 0.5em minus 0.4em\relax PMLR,
  17--23 Jul 2022, pp. 20\,827--20\,840. [Online]. Available:
  \url{https://proceedings.mlr.press/v162/sun22d.html}
\BIBentrySTDinterwordspacing

\bibitem{MTA}
\BIBentryALTinterwordspacing
B.~Duncan, ``Malware traffic analysis,'' 2021. [Online]. Available:
  \url{https://www.malware-traffic-analysis.net/}
\BIBentrySTDinterwordspacing

\bibitem{quicclassificationfewlabels}
S.~Rezaei and X.~Liu, ``How to achieve high classification accuracy with just a
  few labels: {A} semi-supervised approach using sampled packets,''
  \emph{CoRR}, 2018.

\bibitem{maldist_ccnc}
O.~Bader, A.~Lichy, C.~Hajaj, R.~Dubin, and A.~Dvir, ``Maldist: From encrypted
  traffic classification to malware traffic detection and classification,'' in
  \emph{CCNC}.\hskip 1em plus 0.5em minus 0.4em\relax IEEE, 2022.

\bibitem{featuresel}
\BIBentryALTinterwordspacing
J.~Brownlee, ``How to choose a feature selection method for machine learning,''
  2019. [Online]. Available:
  \url{https://machinelearningmastery.com/feature-selection-with-real-and-categorical-data/}
\BIBentrySTDinterwordspacing

\bibitem{featureselection}
\BIBentryALTinterwordspacing
A.~Weizmann and R.~Buskila, ``Feature selection.'' [Online]. Available:
  \url{https://github.com/ComputerScienceMasterStudent/EncryptedTrafficClassification}
\BIBentrySTDinterwordspacing

\bibitem{jeong2020ood}
T.~Jeong and H.~Kim, ``Ood-maml: Meta-learning for few-shot out-of-distribution
  detection and classification,'' \emph{Advances in Neural Information
  Processing Systems}, vol.~33, pp. 3907--3916, 2020.

\bibitem{balltree}
\BIBentryALTinterwordspacing
H.~Mariuse, ``Tree algorithms explained: ball tree algorithm vs. kd tree vs.
  brute force,'' 2020. [Online]. Available:
  \url{https://towardsdatascience.com/tree-algorithms-explained-ball-tree-algorithm-vs-kd-tree-vs-brute-force-9746debcd940}
\BIBentrySTDinterwordspacing

\bibitem{mount1998ann}
D.~M. Mount, ``Ann programming manual,'' Technical report, Dept. of Computer
  Science, U. of Maryland, Tech. Rep., 1998.

\bibitem{annoy2018}
E.~Bernhardsson, ``New approximate nearest neighbor benchmarks,'' 2018.

\bibitem{garcia2010k}
V.~Garcia, E.~Debreuve, F.~Nielsen, and M.~Barlaud, ``K-nearest neighbor
  search: Fast gpu-based implementations and application to high-dimensional
  feature matching,'' in \emph{2010 IEEE International Conference on Image
  Processing}.\hskip 1em plus 0.5em minus 0.4em\relax IEEE, 2010, pp.
  3757--3760.

\bibitem{ramasubramani2020freud}
V.~Ramasubramani, B.~D. Dice, E.~S. Harper, M.~P. Spellings, J.~A. Anderson,
  and S.~C. Glotzer, ``freud: A software suite for high throughput analysis of
  particle simulation data,'' \emph{Computer Physics Communications}, vol. 254,
  p. 107275, 2020.

\bibitem{chen2019robustiq}
W.~Chen, J.~Chen, F.~Zou, Y.-F. Li, P.~Lu, and W.~Zhao, ``Robustiq: A robust
  ann search method for billion-scale similarity search on gpus,'' in
  \emph{Proceedings of the 2019 on international conference on multimedia
  retrieval}, 2019, pp. 132--140.

\bibitem{rahman2022evaluating}
M.~M. Rahman and J.~Te{\v{s}}i{\'c}, ``Evaluating hybrid approximate nearest
  neighbor indexing and search (hannis) for high-dimensional image feature
  search,'' in \emph{2022 IEEE International Conference on Big Data (Big
  Data)}.\hskip 1em plus 0.5em minus 0.4em\relax IEEE, 2022, pp. 6802--6804.

\bibitem{nearpy2013}
O.~Krause-Sparmann, ``Ann search in large, high-dimensional data sets (in
  python),'' 2013.

\bibitem{zhang2019sextant}
Z.~Zhang, X.~Huang, C.~Sun, S.~Zheng, B.~Hu, J.~Varadarajan, Y.~Yin,
  R.~Zimmermann, and G.~Wang, ``Sextant: Grab's scalable in-memory spatial data
  store for real-time k-nearest neighbour search,'' in \emph{2019 20th IEEE
  International Conference on Mobile Data Management (MDM)}.\hskip 1em plus
  0.5em minus 0.4em\relax IEEE, 2019, pp. 243--251.

\bibitem{yan2019k}
D.~Yan, Y.~Wang, J.~Wang, H.~Wang, and Z.~Li, ``K-nearest neighbor search by
  random projection forests,'' \emph{IEEE Transactions on Big Data}, vol.~7,
  no.~1, pp. 147--157, 2019.

\bibitem{voip}
\BIBentryALTinterwordspacing
FCC, ``Voice over internet protocol,'' 2023. [Online]. Available:
  \url{https://www.fcc.gov/general/voice-over-internet-protocol-voip}
\BIBentrySTDinterwordspacing

\bibitem{vod}
\BIBentryALTinterwordspacing
O.~Afilabi, ``What is video on demand (vod) streaming and how does it work?''
  2022. [Online]. Available:
  \url{https://www.makeuseof.com/what-is-video-on-demand-how-it-works/}
\BIBentrySTDinterwordspacing

\end{thebibliography}
\end{document}